\documentclass{ieeeaccess}
\usepackage{cite}
\usepackage{amsmath,amssymb,amsfonts}
\usepackage{algorithmic}
\usepackage{graphicx}
\usepackage{textcomp}
\usepackage{amsmath,amsfonts}
\usepackage{graphicx}
\usepackage{longtable}
\usepackage{multirow}
\usepackage{tabularx}
\usepackage{booktabs}
\usepackage{fixltx2e}
\usepackage{lipsum}
\usepackage{mwe}
\usepackage{dblfloatfix}
\usepackage{hyperref}
\usepackage{subcaption} 
\def\BibTeX{{\rm B\kern-.05em{\sc i\kern-.025em b}\kern-.08em
    T\kern-.1667em\lower.7ex\hbox{E}\kern-.125emX}}
\begin{document}
\history{Preprint submitted to IEEE Access}
\doi{10.1109/ACCESS.2017.DOI}

\title{Comparative analysis of word embeddings in assessing semantic similarity of complex sentences}
\author{\uppercase{Dhivya Chandrasekaran},
\uppercase{and Vijay Mago}}
\address{Lakehead University, Thunder Bay, Ontario, Canada P7B 5E1}

\tfootnote{This work was supported by the Ontario Council on Articulation and Transfer (ONCAT)}

\markboth
{Chandrasekaran \headeretal}
{Chandrasekaran \headeretal}

\corresp{Corresponding author: Dhivya Chandrasekaran (e-mail: dchandra@lakeheadu.ca)}

\begin{abstract}
Semantic textual similarity is one of the open research challenges in the field of Natural Language Processing. Extensive research has been carried out in this field and near-perfect results are achieved by recent transformer-based models in existing benchmark datasets like the STS dataset and the SICK dataset. In this paper, we study the sentences in these datasets and analyze the sensitivity of various word embeddings with respect to the complexity of the sentences. We build a complex sentences dataset comprising of 50 sentence pairs with associated semantic similarity values provided by 15 human annotators. Readability analysis is performed to highlight the increase in complexity of the sentences in the existing benchmark datasets and those in the proposed dataset. Further, we perform a comparative analysis of the performance of various word embeddings and language models on the existing benchmark datasets and the proposed dataset. The results show the increase in complexity of the sentences has a significant impact on the performance of the embedding models resulting in a 10-20\% decrease in Pearson's and Spearman's correlation.
\end{abstract}


\titlepgskip=-15pt

\maketitle

\section{Introduction}
\label{sec:introduction}
\PARstart{O}{ne} of the core components of Natural Language Processing (NLP) is assessing the semantic similarity between text data. The versatility of natural languages has made it a challenging task for researchers to capture the semantics of text data using numerical representations. Measuring the semantics of text data is essential in various NLP tasks like text summarization\cite{mohamed2019srl}, topic modelling\cite{liu2015topical}, text simplification\cite{sulem2018semantic}, machine translation\cite{wieting2019beyond}, question answering tasks\cite{bordes2014question}, information retrieval\cite{kim2017bridging} and so on. Extensive research has been carried out in the past decade in the field of semantic similarity to construct vector representations that preserve the syntactic and semantic properties of words\cite{chandrasekaran2021evolution}. Word embeddings like \textit{word2vec}\cite{mikolov2013efficient} and \textit{GloVe}\cite{pennington2014glove} exploit the principle of the distributional hypothesis \cite{gorman2006scaling} i.e., “similar words occur in similar context”. These methods use the advancements in deep learning techniques to capture the semantics of the words using large text corpora. Recent language models like BERT\cite{devlin2019bert}, RoBERTa\cite{liu2019roberta}, and ALBERT\cite{lan2019albert} use transformers to build vector representations of text data from underlying corpora by traversing through the corpora in both directions. Over the years various benchmark datasets have been used for comparing the performance of models in measuring semantic similarity between text data. Two of the most popular datasets are the STS benchmark dataset\cite{shao2017hcti} and the SICK dataset\cite{marelli2014sick} on which the BERT models have achieved near-perfect results \cite{devlin2019bert}. Analyzing the readability of the sentences in these datasets, we find that the sentences in these datasets have a low readability index which is a measure of complexity of sentences \cite{al2020exploring}. However, various real world applications of semantic similarity involves more complex sentences to be analysed\cite{8685082}. In this paper, a new dataset  with sentences of a higher degree of complexity than the existing datasets is proposed. The dataset comprises 50 pairs of complex sentences, with corresponding similarity scores provided by 15 human annotators. A comparative analysis of various existing text embeddings is performed on two existing benchmark dataset and the proposed complex sentence dataset. The results indicate that the performance of the embedding models decrease significantly with the increase in complexity of the sentences. To the best of our knowledge this research work is the first attempt to analyse the impact of complexity of sentences on the performance of text embedding models.

The remaining of the paper is organised as follows. Section \ref{back} of this paper provides a brief description of the existing research works carried out in the field of semantic similarity. Section \ref{datasets} describes two of the existing benchmark datasets and five different word-embeddings chosen for the comparative analysis. Section \ref{methodology} discusses in detail the methodology adopted to construct the new benchmark dataset and provides a detailed description of the readability analysis that compares the complexity of sentences in the existing datasets to the sentences in the proposed dataset. Section \ref{results} provides a comparative study of the performance of various word embeddings and provides an insight into the inferences made that would guide the future scope of this research. 
\section{Related Work}\label{back}
Similarity between text data does not always attribute to the lexical or syntactic similarity between text. While two sentences that contain exactly the same words may mean something completely different, it is possible that sentences with different sets of words provide the same meaning. Hence while assessing the similarity between text data, it is important to understand the meaning conveyed by the text. The similarity between the meaning of the text is known as semantic textual similarity (STS). For the past three decades, various semantic similarity methods have been proposed to measure semantic similarity. These methods are widely classified as knowledge-based methods and corpus-based methods. The knowledge-based methods use structurally strong ontologies like  Wordnet \cite{miller1995wordnet}, DBPedia\cite{bizer2009dbpedia}, Wikipedia\footnote{\label{fn1}https://www.wikipedia.org/}, Wikitionary\footnote{https://www.wiktionary.org/}, etc. These ontologies are often used like graphs and various edge counting methods, consider the words in the taxonomy as nodes, and calculate the distance between the words using the edges between them. The greater the distance between the words the lower their similarity value\cite{rada1989development}. However, these methods assume that the length of these edges to be similar which is not always the case. Another type of knowledge-based approach, called the feature-based methods, assess the similarity based on features of the words, like their dictionary definition, neighboring concepts, etc. derived from the underlying ontologies\cite{SANCHEZ20127718}. Knowledge-based methods are computationally simple and are efficient in distinguishing the different meanings of words solving the problem of ambiguity with concepts like polysemy and synonymy. However they are heavily dependent on the underlying taxonomies, they do not account for the versatility of natural language, and structured taxonomies for languages other than English are not common\cite{chandrasekaran2021evolution}. Corpus-based semantic similarity methods use statistical principles to capture the contextual semantics of data using large underlying corpora. The principle of distributional hypothesis states that words with similar meanings occur together in documents and this principle forms the basis of most corpus-based methods, while these methods do not take into account the actual meaning of individual words. Word vectors, also called word embeddings, are constructed using corpus-based methods and the similarity is measured based on the angle or distance between these vectors. The dimensionality of these embeddings depends on the size of the corpus. While using significantly large corpora various dimensionality reduction techniques like Singular Value Decomposition (SVD), Principal Component Analysis (PCA), and filtering techniques are used to achieve computational efficiency. These word embeddings are the fundamental components of recent techniques that use the advancements in deep neural networks to achieve a significant increase in performance in semantic similarity tasks. word2vec proposed by Milokov \textit{et al.}\cite{mikolov2013efficient} and GloVe vectors proposed by Pennington \textit{et al.}\cite{pennington2014glove} have proven to be major breakthroughs in the field of semantic similarity and they are two of the most widely used word embeddings to date. In 2019, Delvin \textit{et al.}\cite{devlin2019bert} proposed the Bidirectional Encoder Representation from Transformers (BERT) language model which used transformers to build word embeddings, which were further used for various downstream NLP applications. Variations of the BERT models like, ALBERT\cite{lan2019albert} and RoBERTa\cite{liu2019roberta} were also published in 2019 and have outperformed the existing semantic similarity models achieving state of the art results\cite{qudarsurvey}. Raffel \textit{et al.}\cite{raffel2020exploring} proposed the T5: text-to-text transformer model which used the principle of transfer learning and was trained on a custom-built corpus called “Colossal Clean Crawled Corpus” or C4. This model tied for the first place with ALBERT by achieving a Pearson's correlation of 0.925 on the STS dataset. In the following section, we describe in detail the two most widely used benchmark datasets for assessing the performance of semantic similarity methods, and five of the most popular word embeddings that we have chosen for comparison in this paper.

\section{Datasets and Word embedding models}\label{datasets}
\subsection{Semantic similarity Datasets}
The first and the most widely used word-to-word similarity dataset, the R\&G dataset\cite{rubenstein1965contextual}, was proposed by Rubenstein and Goodenough in 1965 with 65 English noun pairs annotated by 51 native English speakers with similarity values ranging between 0 and 4. Some of the prominent datasets used are compiled in Table \ref{tab:Dataset list} in Appendix \ref{appendix: C}. While many datasets were published over the years as the benchmark for models measuring sentence level semantic similarity, the SICK dataset, and the STS datasets are those that gained significance. For our analyses and comparison, we choose these two datasets owing to their wide usage and popularity.

\subsubsection{SICK Dataset\cite{marelli2014sick}} 
\begin{figure*}[ht]
    \centering
		\includegraphics[width=0.8\textwidth]{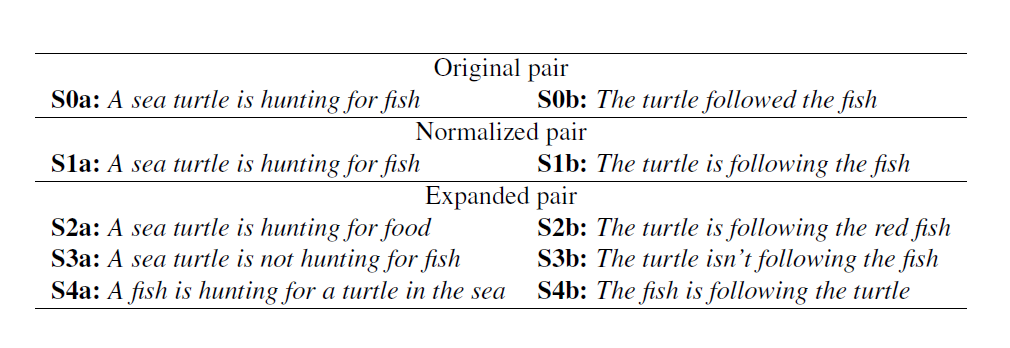}
	\caption{Example of SICK dataset sentence expansion process\cite{marelli2014sick}}
	\label{FIG:1}
\end{figure*}

Marelli \textit{et al.}\cite{marelli2014sick} compiled the SICK dataset for sentence level semantic similarity/relatedness in 2014 composed of 10,000 sentence pairs obtained from the ImageFlickr 8 and MSR-Video descriptions dataset. The sentence pairs were derived from image descriptions by various annotators. 750 random sentence pairs from the two datasets were selected, followed by three steps to obtain the final SICK dataset: sentence normalisation, sentence expansion and sentence pairing. Initially all the sentences were normalised by removing unwanted syntactic or semantic phenomena. This process was carried out by two different annotators and checked for compatibility. In instances of contradiction, a third annotator made the choice by analysing the two alternatives, and if both were correct a random choice was made. From the normalised sentences, 500 pairs were chosen for expansion. The process of expansion involved generating three different versions of the normalised sentence pairs - a similar sentence using meaning preserving transformation, a completely opposite version using negative transformation, and a sentence with the same words but different meaning using random shuffling. FIGURE \ref{FIG:1} shows an example as presented by the authors. Finally, each normalised sentence was paired with all its expanded version, along with some random pairing. A survey was then conducted on Amazon Mechanical Turk where the workers were requested to rank the similarity/relatedness over a scale of 1 to 5 with 1 representing that the sentences were highly dissimilar and 5 representing that the sentences were highly similar. Ten unique responses for each sentence pair were collected and the average of the ten ratings was assigned as the gold standard. Each sentence pair is associated with a relatedness score and a text entailment relation as well. The three entailment relations are NEUTRAL, ENTAILMENT and CONTRADICTION.

\subsubsection{STS Dataset\cite{shao2017hcti}} In order to encourage research in the field of semantic similarity, semantic textual similarity tasks called SemEval have been conducted from 2012. The organizers of the SemEval tasks collected sentences from a wide variety of sources and compiled them to form a benchmark dataset against which the performance of the models submitted by the participants in the task was measured. While the dataset contains different tracks with sentences from different languages, we focus on the English component of the dataset. The English component of the dataset contains 8,295 sentence pairs of which 5,720 are provided as training samples and the remaining 2,575 sentences form the test set. The dataset is built over the years and contains the sentences that were used from 2012 to 2017. Table \ref{tab:tbl1} provides a breakdown of the source of the sentences in the dataset and the year they were appended to form the current version of the dataset. The sentences were annotated using Amazon Mechanical Turk. The quality of the annotators was assessed using the ‘masters’ provided by the platform and five unique annotations were obtained for each sentence pair. The similarity values ranged between the values 0 and 5, with 5 indicating that the sentences are completely similar and 0 indicating the sentences are completely dissimilar. The final similarity score was obtained by taking an average of the five unique responses from the annotators.

\begin{table}[ht]
    
    \caption{STS English language training dataset (2012-2017)\cite{cer2017semeval}}
    \begin{tabularx}{\columnwidth}{p{1cm}|p{2cm}|p{1cm}|p{3.15cm}}
    \toprule
    \textbf{Year} & \textbf{Dataset} & \textbf{Pairs} & \textbf{Source}\\
    \midrule
    2012          & MSRPar           & 1500           & newswire               \\
    2012          & MSRvid           & 1500           & videos                 \\
    2012          & OnWN             & 750            & glosses                \\
    2012          & SMTNews          & 750            & WMT eval.              \\
    2012          & SMTeuroparl      & 750            & WMT eval.              \\
    \midrule
    2013          & HDL              & 750            & newswire               \\
    2013          & FNWN             & 189            & glosses                \\
    2013          & OnWN             & 561            & glosses                \\
    2013          & SMT              & 750            & MT eval.               \\
    \midrule
    2014          & HDL              & 750            & newswire headlines     \\
    2014          & OnWN             & 750            & glosses                \\
    2014          & Deft-forum       & 450            & forum posts            \\
    2014          & Deft-news        & 300            & news summary           \\
    2014          & Images           & 750            & image descriptions     \\
    2014          & Tweet-news       & 750            & tweet-news pairs       \\ \midrule
    2015          & HDL              & 750            & newswire headlines     \\
    2015          & Images           & 750            & image descriptions     \\
    2015          & Ans.-student     & 750            & student answers        \\
    2015          & Ans.-forum       & 375            & Q \& A forum answers   \\
    2015          & Belief           & 375            & committed belief       \\ \midrule
    2016          & HDL              & 249            & newswire headlines     \\
    2016          & Plagiarism       & 230            & short-answers plag.    \\
    2016          & post-editing     & 244            & MT postedits           \\
    2016          & Ans.-Ans         & 254            & Q \& A forum answers   \\
    2016          & Quest.-Quest.    & 209            & Q \& A forum questions \\ \midrule
    2017          & Trail            & 23             & Mixed STS 2016         \\
     \bottomrule
    \end{tabularx}
    \label{tab:tbl1} 
\end{table}
\subsection{Word-embedding models}
Distributed semantic vector representations of words called word embeddings have gained significant attention in the past decade, and a wide variety of word embeddings have been proposed over the years\cite{camacho2018word}. Word embeddings are constructed by analysing the distribution of words in any text data and are well known to capture the semantics of the words thus making them a significant component of a wide variety of semantic similarity algorithms. The \textit{word2vec} model uses neural networks to construct word embeddings and it has been one of the most widely used word-embeddings\cite{camacho2018word}. \textit{GloVe} vectors employ word co-occurrence matrices to identify the distribution of words, which is then statistically used to build word vectors that capture the semantics of the target word with respect to its neighboring words. Pre-trained word embeddings provided by recent transformer based models achieved state of the art results in a wide range of NLP tasks, including semantic similarity. In this section we discuss in detail five of the popular word embeddings that are publicly available.
\subsubsection{word2vec\cite{mikolov2013efficient}} 
\begin{figure}[h]
	\centering\includegraphics[width =2.5in]{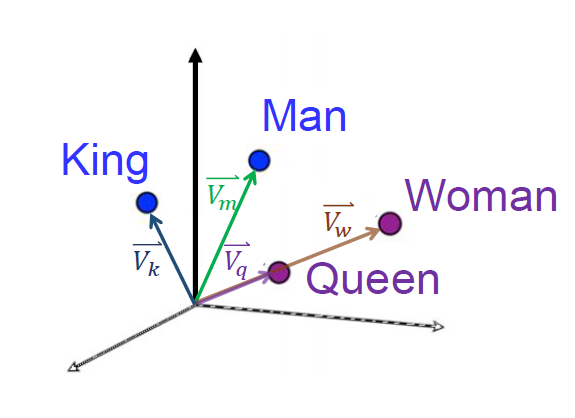}
	\caption{Word vectors in word2vec models[cite]}
	\label{FIG:2}
\end{figure}

Mikolov \textit{et al.} proposed a word embedding called word2vec 2013, using a simple neural network that converted the given input word to a dense vector representation. Two different models of word2vec were proposed namely, the Skip-gram model and the Continuous Bag of Words (CBOW) model. In the skip-gram model, the neural network is optimized to predict a target word given its context words, whereas in the CBOW model, the neural network predicts the neighboring words given a target word. The value vector in the hidden layer of the optimized neural network is used as the vector representation of the word. The number of neurons in the hidden layer of the neural network determines the dimension of the word vector. The models are trained using Google News corpus, which contains 1 million words. The model produced state of the art results in 2013, and is used widely among researchers owing to the simplicity in construction of the neural network. The major claim of the authors was that when simple algebraic operations were performed on these vectors the results were closely related to human understanding. For example, as shown in FIGURE \ref{FIG:2} the difference between $V_k$ (vector representation for the word `king') and $V_m$ (vector representation for the word `man') added to $V_w$ (vector representation for the word `women') provides a vector that is close to vector $V_q$ (vector representation for the word `queen'). This can be mathematically represented as,
\begin{equation}
    Vq \simeq (V_k - V_m) + V_w
\end{equation}
\subsubsection{GloVe\cite{pennington2014glove}} Researchers at Stanford University proposed a vector representation for words using the word to word co-occurrence matrices. Given a corpus, a global co-occurrence matrix is built where each row and column represents the words in the corpus. The underlying principle for the construction of these vectors is that similar words occur together. The model proposed uses ‘log-bilinear regression’ to create a word vector space with substructures that provide meaningful word-vector representations. GloVe vectors were trained on five different corpora like a common web crawled corpus and a Wikipedia data dump resulting in 400,000 unique words to form the co-occurrence matrix. Pretrained word vectors of 3 different dimensions (50, 100 and 300) were released by the authors and they claimed that GloVe vectors outperformed word2vec and achieved the state of the art results in 2014.
\subsubsection{BERT\cite{devlin2019bert}} In 2019, the BERT transformer model surpassed the state of the art results in 11 different NLP tasks, including semantic similarity. BERT uses the transformer model proposed by Vaswani \textit{et al.}\cite{vaswani2017attention}. The BERT models follow two distinct steps to adapt to specific NLP tasks namely, pretraining and fine-tuning. The transformer contains an encoder and decoder module, the encoder containing 6 identical layers stacked above each other. Each layer consists of sublayers comprising of a multi-head attention mechanism followed by a fully connected feed-forward neural network. The decoder is similar to the encoder, with one additional sub-layer of multi-head attention, which captures the attention weights in the output of the encoder. The model is pretrained using the Book corpus\cite{zhu2015aligning} and Wikipedia dump comprising of nearly 3300 million words. Pre-training is carried out with the help of two tasks, namely, `Masked Language Model (MLM)’ and `Next Sentence Prediction (NSP)’. In the first task, random words in the corpus are masked and the model is optimized to predict the masked tokens. In the second task, the model is optimized to predict whether or not a sentence follows another given sentence. The BERT models thus produce bidirectional representations of words taking into consideration the context of the word in both directions. In general, the BERT model is fine-tuned using labeled training data to accommodate a specific NLP task. The model was fine-tuned with STS dataset and achieved state of the art results.
\subsubsection{RoBERTa\cite{liu2019roberta}} Liu \textit{et al.}\cite{liu2019roberta} proposed a robustly optimized version of BERT, by replicating the work of Delvin \textit{et al.}\cite{devlin2019bert} and adding to it an improved training procedure. They added more training data and trained for a longer period of time, achieving state of the results, which proved the BERT architecture was equipped to perform better than many later models, but it was under-trained. While BERT was trained on the Book Corpus and Wikipedia corpus, RoBERTa model was trained on four different corpora namely, the Book Corpus, the Common Crawl News dataset, OpenWebText dataset and the Stories dataset. RoBERTa uses variations of the pretraining tasks used by the BERT model. It uses both static and dynamic masking, and by performing dynamic mask the training data is duplicated ten times thus enabling the model to encounter each masked tokens four times over the forty epoch training. The authors study the effect of `Next Sentence Prediction' task by replacing it with prediction of subsequent segments or sentences in a document and prove that the performance increased by removing the NSP task. The model outperforms the BERT model and achieves state of the art results in 11 NLP tasks including semantic similarity.
\subsubsection{ALBERT\cite{lan2019albert}} One of the major challenges in the BERT model is the time and resource requirement to pretrain a complex transformer model. Lan \textit{et al.}\cite{lan2019albert} proposed a Lite version of BERT by employing two different parameter reduction techniques to aid in scaling the pre-trained models namely, Factorized Embedding Parameterization (FEP) and Cross-layer Parameter Sharing (CPS).  Using FEP, the authors split the vector embedding matrix of the vocabulary into two smaller matrices thus making the size of the vector independent of the hidden layer of the model. Using CPS enables the sharing of parameters across layers thus preventing the increase in number of parameters as the depth of the layers increase. ALBERT also replaces one of the pretraining task, next sentence prediction, in BERT with inter-sentence coherence loss that focuses on the coherence between two consecutive sentences. ALBERT has outperformed all the existing models and hold the highest Pearson's correlation in STS dataset.
\begin{table}[ht]
\caption{STS English language training dataset (2012-2017)\cite{cer2017semeval}}
\begin{tabularx}{\columnwidth}{p{0.5cm}|p{3cm}|p{0.5cm}|p{3.15cm}}
\toprule
\textbf{SNo} & \textbf{Topic} & \textbf{SNo} & \textbf{Topic}\\
\midrule
    1         & Computer science    & 10           & Psuedo code              \\
    2          & Computer program   & 11           & Programming language       \\
    3          & Algorithm        & 12            & Data analytics             \\
    4          & Data structures  & 13            & Computer Security          \\
    5          & Artificial Intelligence  & 14            & Computer virus     \\

    6          & Computer programming  & 15            & Cloud computing     \\
    7          & Operating systems     & 16            & Server                \\
    8          & Database             & 17            & Firewall                \\
    9          & Computer Architecture  & 18            & Outlier.              \\
\bottomrule
\end{tabularx}
\label{tbl2}\\ 
\end{table}

\section{Methodology}\label{methodology}
\begin{figure}[h]
		\includegraphics[width=\columnwidth]{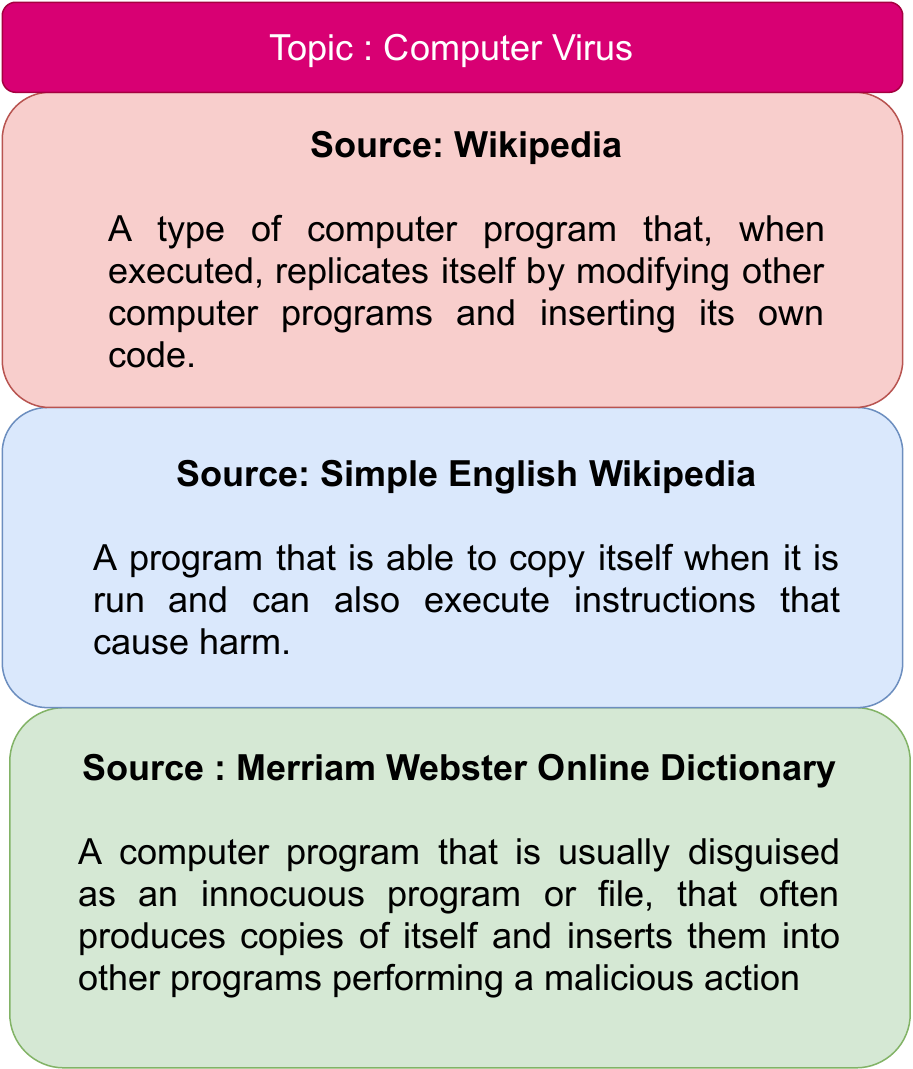}
	\caption{Example of Sentences in the proposed complex sentence dataset.}
	\label{FIG:3}
\end{figure}

In this section, we discuss the methodology followed in building the proposed complex sentence dataset. We follow three steps in the construction of the dataset namely, (1) Selection of terms, (2) Selection of sentences (3) Annotation of similarity values. The dataset comprises 52 unique sentences, which are definitions of widely known topics in the field of computer science. The list of the topics chosen is shown in TABLE \ref{tbl2}. In order to extract sentences with similar meanings we used three different sources namely, the Wikipedia, Simple English Wikipedia\footnote{https://simple.wikipedia.org/wiki/Main\_Page} and the Merriam Webster Online dictionary\footnote{https://www.merriam-webster.com/}. The single sentence definitions of the chosen topics are selected from these sources resulting in 52 unique sentences (two topics have definitions from only two sources.) FIGURE \ref{FIG:3} shows an example of the definitions from various sources. In order to obtain dissimilar sentence pairs the sentences are paired among each other to form 50 final sentence pairs with no two sentences repeated more than twice. These sentence pairs are tabulated in TABLE \ref{tab:datasettable} in Appendix \ref{app1}.
\subsection{Human Annotation}
\subsubsection{Annotation}
The sentence pairs were marked with a `binary annotation' as `Similar' and `Dissimilar' based on whether they are the definition of the same topic to ascertain the ground truth. Then two separate surveys were conducted to obtain the human annotation of the similarity value between the sentence pairs. The initial survey was conducted among 5 graduate students pursuing a Masters in Computer Science. The students were requested to mark the similarity between the provided sentence pairs over a range of values between 0 and 5, where 0 indicates that the sentences are completely dissimilar and 5 indicates that the sentences are completely similar. In order to capture the ratings from a larger audience the survey was further extended to Amazon Mechanical Turk (MTurk) requesting 10 unique responses for each sentence pair. The survey environments of both surveys are depicted in FIGURE \ref{FIG: F5} and FIGURE \ref{FIG: F6} in Appendix \ref{appendix: B}. 
\subsubsection{Validation}
The survey in MTurk was restricted to North America to ensure language expertise and the annotator was required to have US graduate level education for domain expertise. Specific instructions were provided to the workers that sufficient domain knowledge is required to participate in the survey.
However, since the actual qualification of the user cannot be determined or restricted in MTurk the responses of the workers were collected and examined for irregularities by comparing them to the existing `binary annotation' of the dataset mentioned above. If more than 80\% of the similarity values provided by an annotator contradicted the binary annotation it indicates that the annotator does not have enough expertise to respond to the survey, hence, their responses were removed. By repeating this process 10 unique responses for 50 sentence pairs were obtained. Given that the two surveys are conducted over the same range of similarity values their results are combined to obtain 15 unique values for each sentence pair and the similarity score is calculated using the weighted average of the 15 unique responses using the formula below.
\begin{equation}
    S = \frac{\sum_{i=0}^{i=5}(w_i \times s_i)}{\sum_{i=0}^{i=5}w_i}
\end{equation}
where,\newline
$s_i$ represent the values from 0 to 5 respectively and \newline
$w_i$ represent the corresponding weights and are calculated as,
\begin{equation}
    w_i = \left(\frac{\mbox{No. of responses with \textit{i} similarity score}}{\mbox{Total number of responses}}\right)
\end{equation}

 \begin{figure*}[ht]
        \centering
        \begin{subfigure}[b]{\textwidth}
            \centering
            \includegraphics[width=0.7\textwidth]{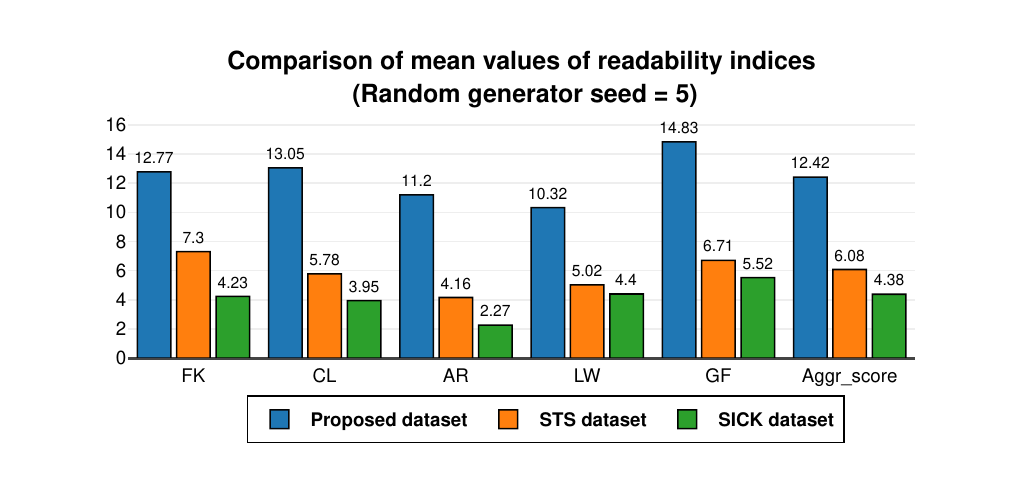}
            \label{Prior-RC1}
        \end{subfigure}
        \begin{subfigure}[b]{\textwidth}  
            \centering 
            \includegraphics[width=0.7\textwidth]{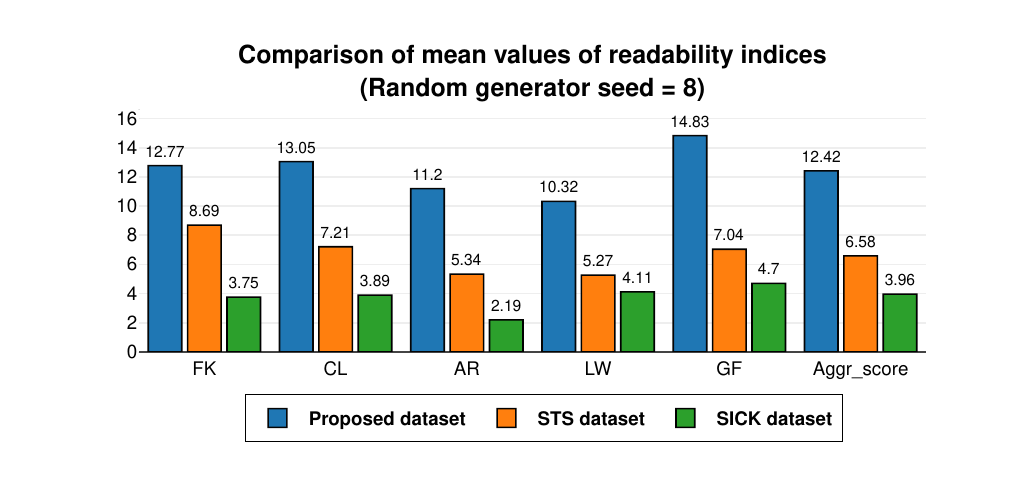}
            \label{Prior-RC2}
        \end{subfigure}
        \begin{subfigure}[b]{\textwidth}
            \centering
            \includegraphics[width=0.7\textwidth]{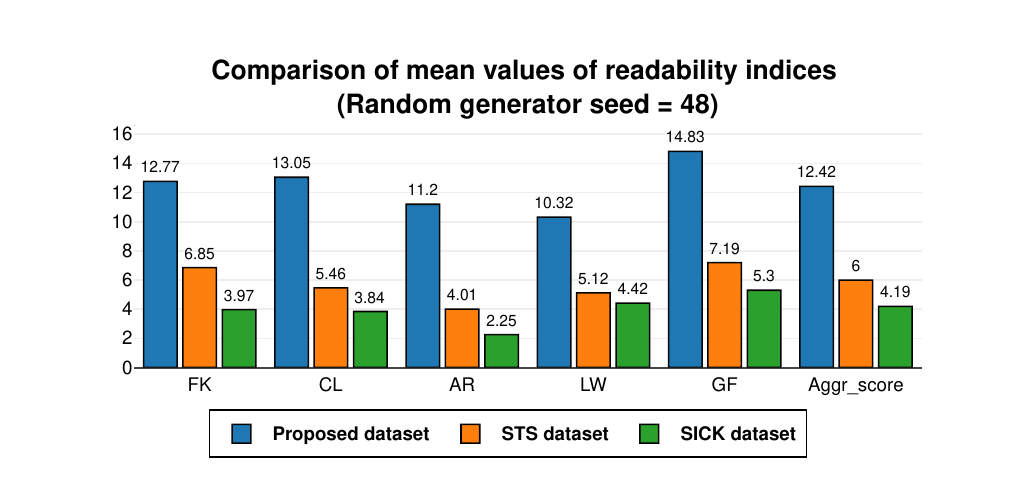}
            \label{Prior-CS1}
        \end{subfigure}
        \hfill
        \caption[Comparison of mean readability indices among three given datasets.]
        {\small Comparison of mean readability indices among three given datasets.} 
        \label{FIG:4}
    \end{figure*}
\subsection{Readability Analysis}
 Readability indices are used by researchers to measure the complexity of text data mostly in text simplification tasks \cite{shardlow2019neural},\cite{al2020exploring}. In order to prove that the sentences chosen for building this dataset are more complex than the existing benchmark datasets a comparative readability analyses is conducted between the two existing benchmark datasets and the proposed dataset. The below-mentioned readability grade-level scores indicate the grade level of education required by the reader to comprehend the given text which in turns reflects the complexity of the sentences. For example, a readability index of 10.4 indicates that a student of grade 10 would be able to read the given text. Various readability indices used and the formula for determining the scores are provided below.
\begin{itemize}
    \item  \textbf{Flesch-Kincaid Grade Level\cite{kincaid1975derivation}}     \begin{multline}
      = 0.39 \left( \frac{total words}{total sentences}\right)\\
      {}+11.8\left(\frac{total syllables}{total words}\right) - 15.59
  \end{multline}
     \item \textbf{Coleman-Liau Index\cite{coleman1975computer}}
    \begin{equation}
    = 0.588L - 0.296S - 15.8
    \end{equation}
    where,\newline
    L denotes the number of characters per 100 words and \newline 
    S denotes the number of sentences per 100 words.
    \item \textbf{Automated readability Index\cite{kincaid1975derivation}}
    \begin{multline}
    = 4.71 \left( \frac{characters}{words}\right)\\
    {}+0.5\left(\frac{words}{sentences}\right) - 21.43
     \end{multline}
    \item \textbf{Linsear Write}\newline
    For each sample of 100 words,\newline
    \begin{equation}
          r = \frac{1\times (Easy words)+3 \times (Hard words)}{\mbox{\textit{No. of sentences in sample}}} 
        \end{equation}
    where,\newline
    Easywords = words with less than 2 syllables and\newline
    Hardwords = words with more than 3 syllables.
    \begin{equation}
        LW = \Bigg\{ \begin{array}{l}\left(\frac{r}{2}\right), \mbox{ if } r>20\\
        
        \left(\frac{r}{2}-1\right), \mbox{ if } r\leq 20
        \end{array}
     \end{equation}
     \item{\textbf{Gunning fog index\cite{gunning1952technique}}}
     \begin{equation}
         = 0.4\Bigg[\left(\frac{words}{sentences}\right)+ 100 \left(\frac{\mbox{\textit{complex words}}}{words}\right)\Bigg]
     \end{equation}
     where,\newline
     \textit{complex words} = words consisting more than or equal to three syllables
     \item\textbf{Text Standard}:\newline
     An aggregated score based on all the above readability indices. 
\end{itemize}

 The STS training dataset contains 10,818 unique sentences and the SICK dataset contains 6,076 unique sentences. On analysing the complexity of these sentences using the above mentioned readability indices we find that 70\% of sentences in STS dataset and 90\% of sentences in SICK dataset have a aggregate readability score below 10, while only 25\% of the sentences in the proposed dataset are below the index 10. This clearly indicates that the two existing datasets have predominantly simpler sentences. In order compare the complexity of the datasets, we select the readability indices of 52 random sentences from the the two benchmark datasets and 52 sentences from the proposed dataset. This process is repeated with three different seeds for random selection. FIGURE \ref{FIG:4} shows the results of the comparison between the mean value of six different readability indices among the three datasets repeated thrice and we can clearly see a significant increase in the complexity of the sentences. The results show us that while sentences in STS dataset and SICK dataset can be interpreted by 6th graders and 4th graders respectively, the proposed dataset requires the knowledge of a 12th grader to comprehend the meaning of the provided sentences clearly indicating the increase in complexity in the sentences of the proposed dataset.
\\
\section{Comparative analysis} \label{results}
\subsection{Experimental setup}
\begin{table}[h]
\caption{Configuration of BERT, RoBERT, and ALBERT models compared in this paper.}
\label{tab:bert}
\centering
\resizebox{\columnwidth}{!}{%
\begin{tabular}{|p{2.4cm}|p{1cm}|p{1cm}|p{1.2cm}|p{1cm}|}
\hline
\textbf{Model Name} & \textbf{Trans former Blocks} & \textbf{Hidden Layers} & \textbf{Attention heads} & \textbf{Total Parameters} \\ \hline
\textbf{BERT\textsubscript{BASE}}                    & 12 & 768  & 12 & 110M \\ 
\textbf{BERT\textsubscript{LARGE}}                   & 24 & 1024 & 16 & 340M \\ 
\textbf{RoBERTa\textsubscript{BASE} and RoBERTa\textsubscript{LARGE}} & 24 & 1024 & 16 & 340M \\ 
\textbf{ALBERT\textsubscript{XLARGE}}               & 24 & 2048 & 32 & 60M  \\ 
\textbf{ALBERT\textsubscript{XXLARGE}}              & 12 & 4096 & 64 & 235M \\ \hline
\end{tabular}%
}
\end{table}

\begin{table*}[h]
 \caption{Pearson's correlation and Spearman's correlation percentages of the word embedding models in STS Benchmark Dataset, SICK dataset, and Proposed dataset.}
 \label{tab:Results}
\centering
\begin{tabular}{|l|l|r|r|r|r|r|r|} 
\hline
\multirow{2}{*}{~ ~\textbf{Supervision}}       & \multirow{2}{*}{~ ~\textbf{Model}}                    & \multicolumn{3}{l|}{\textbf{Pearson's
  Correlation}}                                                   & \multicolumn{3}{l|}{\textbf{Spearman's
  Correlation}}                                                   \\ 
\cline{3-8}
                                      &                                              & \multicolumn{1}{l|}{\textbf{STS}} & \multicolumn{1}{l|}{\textbf{SICK}} & \multicolumn{1}{l|}{\begin{tabular}[c]{@{}c@{}}\textbf{Complex}\\ \textbf{Sentences}\end{tabular}} & \multicolumn{1}{l|}{\textbf{STS}} & \multicolumn{1}{l|}{\textbf{SICK}} & \multicolumn{1}{l|}{\begin{tabular}[c]{@{}c@{}}\textbf{Complex}\\ \textbf{Sentences}\end{tabular}}  \\ 
\hline
\multirow{4}{*}{\textbf{Unsupervised}}         & word2vec                                     & 62.61                           & 72.67                            & 48.30                     & 58.71                           & 62.13                            & 49.04                      \\ 
                                      & GloVe                                        & 45.42                           & 63.53                            & 43.90                     & 46.88                           & 55.59                            & \textbf{50.14}                      \\ 
                                      & word2vec + CS Corpus                         & 56.77                           & 67.75                            & \textbf{52.35}                     & 53.27                           & 59.38                            & 46.79                      \\ 
                                      & GloVe + CS Corpus                            & 41.09                           & 60.44                            & 48.02                     & 42.85                           & 54.20                            & 42.52                      \\ 
\hline
\multirow{6}{*}{\textbf{Partially supervised}} & BERT large + NLI +
  Mean Pooling            & 76.17                           & 73.65                            & 54.78                     & 79.19                           & 73.72                            & 54.47                      \\ 
                                      & BERT base + NLI + Mean Pooling               & 74.11                           & 72.95                            & 50.91                     & 76.97                           & 72.89                            & 47.73                      \\ 
                                      & RoBERTa large + NLI + Mean Pooling         & 76.26                           & 74.35                            & 54.54                     & 78.69                           & 74.01                            & 46.36                      \\ 
                                      & RoBERTa base + NLI + Mean Pooling         & 74.58                           & 76.05                            & 37.74                     & 77.09                           & 74.44                            & 39.48                      \\ 
                                      & ALBERT xxlarge + NLI + Mean Pooling         & 78.56                           & 77.78                            & ~45.44                    & ~79.55                          & ~75.37                           & 41.33                      \\ 
                                      & ALBERT xlarge + NLI + Mean Pooling           & ~77.68                          & 73.90                            & \textbf{54.90}                     & 80.12                           & ~74.78                           & ~\textbf{57.68}                     \\
\hline
\multirow{6}{*}{\textbf{Supervised} ~ ~ ~ ~ ~} & BERT large + NLI +
  STSB+ Mean Pooling      & 84.63                           & 80.97                            & 64.69                     & 85.25                           & 78.48                            & 60.48                      \\ 
                                      & BERT base + NLI
  +~ STSB+ Mean Pooling      & 84.18                           & 82.16                            & 67.79                     & 85.04                           & 78.43                            & 64.02                      \\ 
                                      & RoBERTa large + NLI
  +~ STSB+ Mean Pooling  & 85.54                           & 82.32                            & 71.98                     & 86.42                           & 78.40                            & 66.78                      \\ 
                                      & RoBERTa base + NLI
  +~ STSB+ Mean Pooling   & 84.26                           & 81.78                            & 50.57                     & 85.26                           & 77.43                            & 46.46                      \\ 
                                      & ALBERT xxlarge + NLI
  + STSB + Mean Pooling & 92.58                           & ~85.33                           & \textbf{73.12}                     & ~90.22                          & ~80.03                           & ~\textbf{67.54}                     \\ 
                                      & ALBERT xlarge + NLI +
  STSB + Mean Pooling  & ~91.59                          & ~83.28                           & ~71.45                    & ~90.73                          & 81.52                            & ~66.97                     \\
\hline
\end{tabular}
\end{table*}
We perform a comparative analysis between the five chosen word embeddings, to assess the impact of the complexity of the sentences. We compare their performance across the two benchmark datasets and the proposed dataset. For word embedding models the sentence vectors are formed by simply taking the mean of the word vectors. In transformer based models a mean pooling is added to the model to form the sentence vectors. 
Initially we replicate the results provided by the authors of each model using the STS dataset, and use the same models to measure the similarity scores for both SICK dataset and the proposed complex sentence dataset. In the first model, the pre-trained \textit{word2vec} model, trained on the Google News dataset containing 300 dimensional vectors with a vocabulary of 3M words provided by $gensim$ python library is used for building the word vectors. Then, the 300 dimensional \textit{GloVe} vectors pre-trained on a Wikipedia dump corpus with 6B tokens provided by the $gensim$ library is used to build the word vectors. However, since the proposed sentences are definitions of computer science related topics, we use a specific corpus - the computer science corpus to train the word2vec and GloVe models initialized with the pretrained weights using transfer learning. The PetScan\footnote{https://petscan.wmflabs.org/} tool that traverses through the Wikipedia categories and subcategories based on provided keywords is used for building the corpus. Using ‘computer\_science’ as the category and a depth ‘2’ which indicates the level of subcategories to be included, the `computer science corpus' containing 4M tokens is used to train both the word2vec and GloVe models. Since the different BERT models do not provide explicit word embeddings, we use Sentence-BERT framework proposed by Reimer’s \textit{et al.}\cite{reimers-2019-sentence-bert} to compare their performance. The various BERT models selected for comparison and their configuration is listed above in TABLE \ref{tab:bert}. We use the SentenceTransformer python library to initialize the model with the weights of pretrained BERT models followed by a mean pooling layer to form the sentence vectors. In order to factor in the effect of ‘fine-tuning’ one of the prominent characteristic of transformer based models we finetune the BERT models, with the AllNLI dataset. To estimate the impact of supervised learning in the quality of the word vectors provided by transformer based models we experiment with the BERT models fine tuned with both AllNLI dataset and the STS dataset on all three datasets included in the comparison. While the T5 model proposed by Raffel \textit{et al.}\cite{raffel2020exploring} is one of the models that achieved the best performance in the STS dataset, due to computational constraints we were not able to replicate the model hence it is not included in our comparison. 

\subsection{Results and Discussion}
The correlation between the similarity scores provided by human annotators and the similarity scores calculated by the models is used as the measure to estimate the performance of the word embedding models. Since both Pearson's correlation and Spearman's correlation are used by the authors of the models chosen for comparison, we depict our results using both the correlations. The results are categorized as three different sections based on the level of supervision used for the models. The first section records the performance of the unsupervised word embedding models, the word2vec and GloVe and their variations trained on the `computer science corpus'. The second section shows the performance of the two variants of BERT, RoBERTa and ALBERT models, fine tuned using the AllNLI dataset (Natural Language Inference dataset closely similarity to semantic similarity tasks), we consider this as partial supervision since the text entailment tasks are similar to semantic similarity tasks, but the model has not been trained on sentences from any of the three datasets used for comparison. The third section comprises the results of the same models trained (fine-tuned) using both AllNLI and the STS dataset. The results are provided in TABLE \ref{tab:Results}. \par
The performance of the word-embeddings decrease considerably when tested on the proposed dataset, indicating that the increase in complexity of the text has a significant and inverse effect on the performance of widely used text embedding models. While ALBERT-xxlarge model achieves the best performance in the proposed dataset with 73.67\% Pearson's Correlation and 67.54\% Spearman's correlation, it is clear that these results are sub-par in comparison to the 92.58\% correlation achieved in the existing benchmark dataset. It is important to note that, though the sentences in the proposed dataset are definitions of topics from a particular domain, they are derived from sources that are commonly used by everyone. Hence these sentences are comparatively simpler than the sentences in scientific or academic articles. Hence, it is evident that there is an imminent need to explore venues to improve the quality of these word embeddings to capture the semantics in complex documents. We also see that the BERT models fine-tuned to a specific NLP task like semantic similarity do not perform as effectively in the proposed dataset indicating the impact of the training data used in the fine-tuning process. Researchers have highlighted the need for building better datasets to ensure the robustness of the embedding models. Rogers \textit{et al.}\cite{rogers2020primer} mentions that ``{\it As with any optimization method, if there is a shortcut in the data, we have no reason to expect BERT to not learn it. But harder datasets that cannot be resolved with shallow heuristics are unlikely to emerge if their development is not as valued as modeling work.}". In future, understanding the need for versatile and harder datasets we intend to repeat the process and add more sentences to the dataset thus building a large complex sentence dataset which can in turn be used by transformer-based models for fine tuning.

\section{Conclusion}
Measuring semantic similarity between text data has been one of the most challenging tasks in the field of Natural Language Processing. Various word embedding models have been proposed over the years to capture the semantics of the words in numeric representations. In this article a new dataset with sentences that are more complex than the existing benchmark datasets is built and a read-ability analysis is performed to prove the increase in complexity. A comparative analysis of various word embedding and language models is performed across the two existing benchmark datasets i.e., the STS dataset and SICK dataset and the proposed dataset. The results show that while the models achieve near perfect results in the existing benchmark datasets the correlation drops significantly in the proposed dataset, indicating the impact of increase in complexity of the sentences on the performance of the existing models.
\ifCLASSOPTIONcompsoc
  \section*{Acknowledgments}
\else
  \section*{Acknowledgment}
\fi
The authors would like to extend our gratitude to the research team in the DaTALab at Lakehead University for their support, Punardeep Sikka and Arunim Garg for their feedback and revisions on this publication. We would also like to thank Lakehead University, CASES and the Ontario Council for Articulation and Transfer, without their support this research would not have been possible.
\bibliographystyle{IEEEtran}
\bibliography{references}

\begin{thebibliography}{10}
\providecommand{\url}[1]{#1}
\csname url@samestyle\endcsname
\providecommand{\newblock}{\relax}
\providecommand{\bibinfo}[2]{#2}
\providecommand{\BIBentrySTDinterwordspacing}{\spaceskip=0pt\relax}
\providecommand{\BIBentryALTinterwordstretchfactor}{4}
\providecommand{\BIBentryALTinterwordspacing}{\spaceskip=\fontdimen2\font plus
\BIBentryALTinterwordstretchfactor\fontdimen3\font minus
  \fontdimen4\font\relax}
\providecommand{\BIBforeignlanguage}[2]{{%
\expandafter\ifx\csname l@#1\endcsname\relax
\typeout{** WARNING: IEEEtran.bst: No hyphenation pattern has been}%
\typeout{** loaded for the language `#1'. Using the pattern for}%
\typeout{** the default language instead.}%
\else
\language=\csname l@#1\endcsname
\fi
#2}}
\providecommand{\BIBdecl}{\relax}
\BIBdecl

\bibitem{mohamed2019srl}
M.~Mohamed and M.~Oussalah, ``{SRL-ESA-T}ext{S}um: A text summarization
  approach based on semantic role labeling and explicit semantic analysis,''
  \emph{Information Processing \& Management}, vol.~56, no.~4, pp. 1356--1372,
  2019.

\bibitem{liu2015topical}
Y.~Liu, Z.~Liu, T.-S. Chua, and M.~Sun, ``Topical word embeddings,'' in
  \emph{Twenty-ninth AAAI conference on artificial intelligence}.\hskip 1em
  plus 0.5em minus 0.4em\relax Citeseer, 2015.

\bibitem{sulem2018semantic}
E.~Sulem, O.~Abend, and A.~Rappoport, ``Semantic structural evaluation for text
  simplification,'' in \emph{Proceedings of NAACL-HLT}, 2018, pp. 685--696.

\bibitem{wieting2019beyond}
J.~Wieting, T.~Berg-Kirkpatrick, K.~Gimpel, and G.~Neubig, ``Beyond bleu:
  Training neural machine translation with semantic similarity,'' in
  \emph{Proceedings of the 57th Annual Meeting of the Association for
  Computational Linguistics}, 2019, pp. 4344--4355.

\bibitem{bordes2014question}
A.~Bordes, S.~Chopra, and J.~Weston, ``Question answering with subgraph
  embeddings,'' in \emph{Proceedings of the 2014 Conference on Empirical
  Methods in Natural Language Processing (EMNLP)}, 2014, pp. 615--620.

\bibitem{kim2017bridging}
S.~Kim, N.~Fiorini, W.~J. Wilbur, and Z.~Lu, ``Bridging the gap: Incorporating
  a semantic similarity measure for effectively mapping pubmed queries to
  documents,'' \emph{Journal of biomedical informatics}, vol.~75, pp. 122--127,
  2017.

\bibitem{chandrasekaran2021evolution}
D.~Chandrasekaran and V.~Mago, ``Evolution of {S}emantic {S}imilarity-{A}
  {S}urvey,'' \emph{ACM Computing Surveys (CSUR)}, vol.~54, no.~2, pp. 1--37,
  2021.

\bibitem{mikolov2013efficient}
T.~Mikolov, K.~Chen, G.~Corrado, and J.~Dean, ``Efficient estimation of word
  representations in vector space,'' \emph{arXiv preprint arXiv:1301.3781},
  2013.

\bibitem{pennington2014glove}
J.~Pennington, R.~Socher, and C.~D. Manning, ``Glove: Global vectors for word
  representation,'' in \emph{Proceedings of the 2014 conference on empirical
  methods in natural language processing (EMNLP)}, 2014, pp. 1532--1543.

\bibitem{gorman2006scaling}
J.~Gorman and J.~R. Curran, ``Scaling distributional similarity to large
  corpora,'' in \emph{Proceedings of the 21st International Conference on
  Computational Linguistics and 44th Annual Meeting of the Association for
  Computational Linguistics}, 2006, pp. 361--368.

\bibitem{devlin2019bert}
J.~Devlin, M.-W. Chang, K.~Lee, and K.~Toutanova, ``Bert: Pre-training of deep
  bidirectional transformers for language understanding,'' in \emph{Proceedings
  of the 2019 Conference of the North American Chapter of the Association for
  Computational Linguistics: Human Language Technologies, Volume 1 (Long and
  Short Papers)}, 2019, pp. 4171--4186.

\bibitem{liu2019roberta}
Y.~Liu, M.~Ott, N.~Goyal, J.~Du, M.~Joshi, D.~Chen, O.~Levy, M.~Lewis,
  L.~Zettlemoyer, and V.~Stoyanov, ``Ro{BERT}a: A robustly optimized {BERT}
  pretraining approach,'' \emph{arXiv preprint arXiv:1907.11692}, 2019.

\bibitem{lan2019albert}
Z.~Lan, M.~Chen, S.~Goodman, K.~Gimpel, P.~Sharma, and R.~Soricut, ``Albert: A
  lite bert for self-supervised learning of language representations,'' in
  \emph{International Conference on Learning Representations}, 2019.

\bibitem{shao2017hcti}
Y.~Shao, ``Hcti at semeval-2017 task 1: Use convolutional neural network to
  evaluate semantic textual similarity,'' in \emph{Proceedings of the 11th
  International Workshop on Semantic Evaluation (SemEval-2017)}, 2017, pp.
  130--133.

\bibitem{marelli2014sick}
M.~Marelli, S.~Menini, M.~Baroni, L.~Bentivogli, R.~Bernardi, R.~Zamparelli
  \emph{et~al.}, ``A {SICK} cure for the evaluation of compositional
  distributional semantic models.'' in \emph{LREC}, 2014, pp. 216--223.

\bibitem{al2020exploring}
J.~Al~Qundus, A.~Paschke, S.~Gupta, A.~M. Alzouby, and M.~Yousef, ``Exploring
  the impact of short-text complexity and structure on its quality in social
  media,'' \emph{Journal of Enterprise Information Management}, 2020.

\bibitem{8685082}
A.~{Heppner}, A.~{Pawar}, D.~{Kivi}, and V.~{Mago}, ``Automating articulation:
  Applying natural language processing to post-secondary credit transfer,''
  \emph{IEEE Access}, vol.~7, pp. 48\,295--48\,306, 2019.

\bibitem{miller1995wordnet}
G.~A. Miller, ``Wordnet: A lexical database for {E}nglish,''
  \emph{Communications of the ACM}, vol.~38, no.~11, pp. 39--41, 1995.

\bibitem{bizer2009dbpedia}
C.~Bizer, J.~Lehmann, G.~Kobilarov, S.~Auer, C.~Becker, R.~Cyganiak, and
  S.~Hellmann, ``{DB}pedia-a crystallization point for the web of data,''
  \emph{Journal of web semantics}, vol.~7, no.~3, pp. 154--165, 2009.

\bibitem{rada1989development}
R.~Rada, H.~Mili, E.~Bicknell, and M.~Blettner, ``Development and application
  of a metric on semantic nets,'' \emph{IEEE transactions on systems, man, and
  cybernetics}, vol.~19, no.~1, pp. 17--30, 1989.

\bibitem{SANCHEZ20127718}
\BIBentryALTinterwordspacing
D.~Sánchez, M.~Batet, D.~Isern, and A.~Valls, ``Ontology-based semantic
  similarity: A new feature-based approach,'' \emph{Expert Systems with
  Applications}, vol.~39, no.~9, pp. 7718 -- 7728, 2012. [Online]. Available:
  \url{http://www.sciencedirect.com/science/article/pii/S0957417412000954}
\BIBentrySTDinterwordspacing

\bibitem{qudarsurvey}
M.~M.~A. QUDAR and V.~MAGO, ``A survey on language models,'' 2020.

\bibitem{raffel2020exploring}
C.~Raffel, N.~Shazeer, A.~Roberts, K.~Lee, S.~Narang, M.~Matena, Y.~Zhou,
  W.~Li, and P.~J. Liu, ``Exploring the limits of transfer learning with a
  unified text-to-text transformer,'' \emph{Journal of Machine Learning
  Research}, vol.~21, no. 140, pp. 1--67, 2020.

\bibitem{rubenstein1965contextual}
H.~Rubenstein and J.~B. Goodenough, ``Contextual correlates of synonymy,''
  \emph{Communications of the ACM}, vol.~8, no.~10, pp. 627--633, 1965.

\bibitem{cer2017semeval}
D.~Cer, M.~Diab, E.~Agirre, I.~Lopez-Gazpio, and L.~Specia, ``Semeval-2017 task
  1: Semantic textual similarity multilingual and crosslingual focused
  evaluation,'' in \emph{Proceedings of the 11th International Workshop on
  Semantic Evaluation (SemEval-2017)}, 2017, pp. 1--14.

\bibitem{camacho2018word}
J.~Camacho-Collados and M.~T. Pilehvar, ``From word to sense embeddings: A
  survey on vector representations of meaning,'' \emph{Journal of Artificial
  Intelligence Research}, vol.~63, pp. 743--788, 2018.

\bibitem{vaswani2017attention}
A.~Vaswani, N.~Shazeer, N.~Parmar, J.~Uszkoreit, L.~Jones, A.~N. Gomez,
  L.~Kaiser, and I.~Polosukhin, ``Attention is all you need,'' in \emph{NIPS},
  2017.

\bibitem{zhu2015aligning}
Y.~Zhu, R.~Kiros, R.~Zemel, R.~Salakhutdinov, R.~Urtasun, A.~Torralba, and
  S.~Fidler, ``Aligning books and movies: Towards story-like visual
  explanations by watching movies and reading books,'' in \emph{Proceedings of
  the IEEE international conference on computer vision}, 2015, pp. 19--27.

\bibitem{shardlow2019neural}
M.~Shardlow and R.~Nawaz, ``Neural text simplification of clinical letters with
  a domain specific phrase table,'' in \emph{Proceedings of the 57th Annual
  Meeting of the Association for Computational Linguistics}, 2019, pp.
  380--389.

\bibitem{kincaid1975derivation}
J.~P. Kincaid, R.~P. Fishburne~Jr, R.~L. Rogers, and B.~S. Chissom,
  ``Derivation of new readability formulas (automated readability index, fog
  count and flesch reading ease formula) for navy enlisted personnel,'' Naval
  Technical Training Command Millington TN Research Branch, Tech. Rep., 1975.

\bibitem{coleman1975computer}
M.~Coleman and T.~L. Liau, ``A computer readability formula designed for
  machine scoring.'' \emph{Journal of Applied Psychology}, vol.~60, no.~2, p.
  283, 1975.

\bibitem{gunning1952technique}
R.~Gunning \emph{et~al.}, ``Technique of clear writing,'' 1952.

\bibitem{reimers-2019-sentence-bert}
N.~Reimers, I.~Gurevych, N.~Reimers, I.~Gurevych, N.~Thakur, N.~Reimers,
  J.~Daxenberger, I.~Gurevych, N.~Reimers, I.~Gurevych \emph{et~al.},
  ``Sentence-bert: Sentence embeddings using siamese bert-networks,'' in
  \emph{Proceedings of the 2019 Conference on Empirical Methods in Natural
  Language Processing}.\hskip 1em plus 0.5em minus 0.4em\relax Association for
  Computational Linguistics, 2019.

\bibitem{rogers2020primer}
A.~Rogers, O.~Kovaleva, and A.~Rumshisky, ``A primer in {BERT}ology: What we
  know about how {BERT} works,'' \emph{Transactions of the Association for
  Computational Linguistics}, vol.~8, pp. 842--866, 2020.

\bibitem{miller1991contextual}
G.~A. Miller and W.~G. Charles, ``Contextual correlates of semantic
  similarity,'' \emph{Language and cognitive processes}, vol.~6, no.~1, pp.
  1--28, 1991.

\bibitem{finkelstein2001placing}
L.~Finkelstein, E.~Gabrilovich, Y.~Matias, E.~Rivlin, Z.~Solan, G.~Wolfman, and
  E.~Ruppin, ``Placing search in context: The concept revisited,'' in
  \emph{Proceedings of the 10th international conference on World Wide Web},
  2001, pp. 406--414.

\bibitem{li2006sentence}
Y.~Li, D.~McLean, Z.~A. Bandar, J.~D. O'shea, and K.~Crockett, ``Sentence
  similarity based on semantic nets and corpus statistics,'' \emph{IEEE
  transactions on knowledge and data engineering}, vol.~18, no.~8, pp.
  1138--1150, 2006.

\bibitem{pedersen2007measures}
T.~Pedersen, S.~V. Pakhomov, S.~Patwardhan, and C.~G. Chute, ``Measures of
  semantic similarity and relatedness in the biomedical domain,'' \emph{Journal
  of biomedical informatics}, vol.~40, no.~3, pp. 288--299, 2007.

\bibitem{agirre2009study}
E.~Agirre, E.~Alfonseca, K.~Hall, J.~Kravalov{\'a}, M.~Pasca, and A.~Soroa, ``A
  study on similarity and relatedness using distributional and wordnet-based
  approaches,'' in \emph{Proceedings of Human Language Technologies: The 2009
  Annual Conference of the North American Chapter of the Association for
  Computational Linguistics}, 2009, pp. 19--27.

\bibitem{agirre2012semeval}
E.~Agirre, D.~Cer, M.~Diab, and A.~Gonzalez-Agirre, ``Semeval-2012 task 6: A
  pilot on semantic textual similarity,'' in \emph{* SEM 2012: The First Joint
  Conference on Lexical and Computational Semantics--Volume 1: Proceedings of
  the main conference and the shared task, and Volume 2: Proceedings of the
  Sixth International Workshop on Semantic Evaluation (SemEval 2012)}, 2012,
  pp. 385--393.

\bibitem{agirre2013sem}
E.~Agirre, D.~Cer, M.~Diab, A.~Gonzalez-Agirre, and W.~Guo, ``* sem 2013 shared
  task: Semantic textual similarity,'' in \emph{Second Joint Conference on
  Lexical and Computational Semantics (* SEM), Volume 1: Proceedings of the
  Main Conference and the Shared Task: Semantic Textual Similarity}, 2013, pp.
  32--43.

\bibitem{li2013computing}
P.~Li, H.~Wang, K.~Q. Zhu, Z.~Wang, and X.~Wu, ``Computing term similarity by
  large probabilistic isa knowledge,'' in \emph{Proceedings of the 22nd ACM
  international conference on Information \& Knowledge Management}, 2013, pp.
  1401--1410.

\bibitem{agirre2014semeval}
E.~Agirre, C.~Banea, C.~Cardie, D.~Cer, M.~Diab, A.~Gonzalez-Agirre, W.~Guo,
  R.~Mihalcea, G.~Rigau, and J.~Wiebe, ``Semeval-2014 task 10: Multilingual
  semantic textual similarity,'' in \emph{Proceedings of the 8th international
  workshop on semantic evaluation (SemEval 2014)}, 2014, pp. 81--91.

\bibitem{silberer2014learning}
C.~Silberer and M.~Lapata, ``Learning grounded meaning representations with
  autoencoders,'' in \emph{Proceedings of the 52nd Annual Meeting of the
  Association for Computational Linguistics (Volume 1: Long Papers)}, 2014, pp.
  721--732.

\bibitem{hill2015simlex}
F.~Hill, R.~Reichart, and A.~Korhonen, ``Simlex-999: Evaluating semantic models
  with (genuine) similarity estimation,'' \emph{Computational Linguistics},
  vol.~41, no.~4, pp. 665--695, 2015.

\bibitem{agirre2015semeval}
E.~Agirre, C.~Banea, C.~Cardie, D.~Cer, M.~Diab, A.~Gonzalez-Agirre, W.~Guo,
  I.~Lopez-Gazpio, M.~Maritxalar, R.~Mihalcea \emph{et~al.}, ``Semeval-2015
  task 2: Semantic textual similarity, {E}nglish, {S}panish and pilot on
  interpretability,'' in \emph{Proceedings of the 9th international workshop on
  semantic evaluation (SemEval 2015)}, 2015, pp. 252--263.

\bibitem{gerz2016simverb}
D.~Gerz, I.~Vuli{\'c}, F.~Hill, R.~Reichart, and A.~Korhonen, ``Simverb-3500: A
  large-scale evaluation set of verb similarity,'' in \emph{Proceedings of the
  2016 Conference on Empirical Methods in Natural Language Processing}, 2016,
  pp. 2173--2182.

\bibitem{agirre2016semeval}
E.~Agirre, C.~Banea, D.~Cer, M.~Diab, A.~Gonzalez~Agirre, R.~Mihalcea,
  G.~Rigau~Claramunt, and J.~Wiebe, ``Semeval-2016 task 1: Semantic textual
  similarity, monolingual and cross-lingual evaluation,'' in
  \emph{SemEval-2016. 10th International Workshop on Semantic Evaluation; 2016
  Jun 16-17; San Diego, CA. Stroudsburg (PA): ACL; 2016. p. 497-511.}\hskip 1em
  plus 0.5em minus 0.4em\relax ACL (Association for Computational Linguistics),
  2016.

\bibitem{pilehvar2019wic}
M.~T. Pilehvar and J.~Camacho-Collados, ``Wic: the word-in-context dataset for
  evaluating context-sensitive meaning representations,'' in \emph{Proceedings
  of the 2019 Conference of the North American Chapter of the Association for
  Computational Linguistics: Human Language Technologies, Volume 1 (Long and
  Short Papers)}, 2019, pp. 1267--1273.

\end{thebibliography}
\begin{IEEEbiography}[{\includegraphics[width=1in,height=1.25in,clip,keepaspectratio]{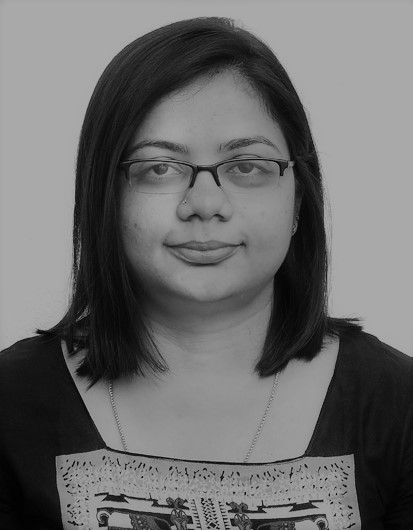}}]{Dhivya Chandrasekaran}
has obtained the Masters degree in Computer Science (AI specialization) at Lakehead University. She received her BEngg degree from Anna University, India. Her research interests include Natural Language Processing, Deep Learning and Aritificial Intelligence. She is a research assistant at DaTALab at Lakehead University.
\end{IEEEbiography}
\begin{IEEEbiography}[{\includegraphics[width=1in,height=1.25in,clip,keepaspectratio]{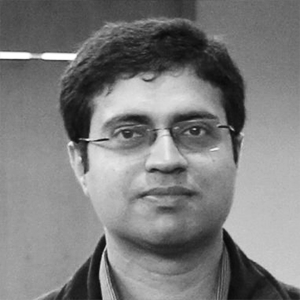}}]{Vijay Mago} received the Ph.D. degree in computer science from Panjab University, India, in 2010. In 2011, he joined the Modeling of Complex Social Systems Program at the IRMACS Centre of Simon Fraser University. He is currently Chair and Associate Professor with the Department of Computer Science, Lakehead University, Thunder Bay, ON, Canada, where he teaches and conducts research in areas, including big data analytics, machine learning, natural language processing, artificial intelligence, medical decision making, and Bayesian intelligence. In 2017, he joined the Technical Investment Strategy Advisory Committee Meeting for Compute Ontario. He has published extensively on new methodologies based on soft computing and artificial intelligence techniques to tackle complex systemic problems, such as homelessness, obesity, and crime. He serves as an Associate Editor for IEEE Access and BMC Medical Informatics and Decision Making and as a Co-Editor for the Journal of Intelligent Systems.
\end{IEEEbiography}
\appendices
\onecolumn
\section{Complex Sentence Dataset}\label{app1}
\begin{footnotesize}
\begin{longtable}[c]{|p{0.5cm}|p{6.5cm}|p{6.5cm}|p{2cm}|} \hline
      \textbf{SNo} &
      \textbf{Sentence\_1} &
      \textbf{Sentence\_2} &
      \textbf{Final Similarity out of 5} \\ \hline
    \endfirsthead
    \multicolumn{4}{c}%
    {{\bfseries Table \thetable\ continued from previous page}} \\
    \hline
    \textbf{SNo} &
      \textbf{Sentence\_1} &
      \textbf{Sentence\_2} &
      \textbf{Final Similarity out of 5} \\ \hline
    \endhead
    1 &
      The study of computation and information. &
      The study of manipulating, managing, transforming   and encoding information. &
      3.60 \\ \hline
    2 &
      The study of computation and information. &
      A branch of science that deals with the theory of computation or the   design of computers &
      3.67 \\ \hline
    3 &
      A program can be a plan of how to do something &
      A branch of science that deals with the theory of computation or the   design of computers &
      0.87 \\ \hline
    4 &
      A finite sequence of well-defined, computer-implementable instructions,   typically to solve a class of problems or to perform a computation &
      A specific set of instructions or steps on how to complete a task. &
      4.00 \\ \hline
    5 &
      A finite sequence of well-defined, computer-implementable instructions,   typically to solve a class of problems or to perform a computation &
      A procedure for solving a mathematical problem in a finite number of   steps that frequently involves repetition of an operation &
      3.20 \\ \hline
    6 &
      A data organization, management, and storage format that   enables efficient access and modification. &
      A specific set of instructions or steps on how to complete a task. &
      0.87 \\ \hline
    7 &
      A set of instructions used to control the behavior of a machine. &
      A program can be a plan of how to do something &
      2.13 \\ \hline
    8 &
      A set of instructions used to control the behavior of a machine. &
      A sequence of coded instructions that can be inserted into a mechanism &
      3.27 \\ \hline
    9 &
      The study of manipulating, managing, transforming   and encoding information. &
      A sequence of coded instructions that can be inserted into a mechanism &
      1.87 \\ \hline
    10 &
      A data organization, management, and storage format that   enables efficient access and modification. &
      The organization and implementation of values and information. &
      2.47 \\ \hline
    11 &
      A procedure for solving a mathematical problem in a finite number of   steps that frequently involves repetition of an operation &
      Various methods or formats for organizing data in a computer &
      0.60 \\ \hline
    12 &
      The organization and implementation of values and information. &
      Various methods or formats for organizing data in a computer &
      2.53 \\ \hline
    13 &
      The intelligence demonstrated by machines, unlike   the natural intelligence displayed   by humans and animals. &
      The ability of a computer program or a machine to think and learn. &
      3.67 \\ \hline
    14 &
      The intelligence demonstrated by machines, unlike   the natural intelligence displayed   by humans and animals. &
      A branch of computer science dealing with the simulation of intelligent   behavior in computers &
      3.07 \\ \hline
    15 &
      The ability of a computer program or a machine to think and learn. &
      A branch of computer science dealing with the simulation of intelligent   behavior in computers &
      3.67 \\ \hline
    16 &
      The process of designing and building an executable computer   program to accomplish a specific computing result. &
      A software that controls the operation of a computer and directs the   processing of programs &
      2.40 \\ \hline
    17 &
      The process of designing and building an executable computer   program to accomplish a specific computing result. &
      The process of preparing an instructional program for a device &
      3.00 \\ \hline
    18 &
      The process of telling a computer to do certain things by giving it   instructions. &
      The process of preparing an instructional program for a device &
      3.40 \\ \hline
    19 &
      A system software that manages computer hardware, software resources, and   provides common services for computer programs. &
      A group of computer programs, including device drivers, kernels, and   other software that lets people interact with a computer. &
      2.80 \\ \hline
    20 &
      A system software that manages computer hardware, software resources, and   provides common services for computer programs. &
      The process of telling a computer to do certain things by giving it   instructions. &
      1.73 \\ \hline
    21 &
      A group of computer programs, including device drivers, kernels, and   other software that lets people interact with a computer. &
      A software that controls the operation of a computer and directs the   processing of programs &
      3.27 \\ \hline
    22 &
      An organized collection of data, generally stored and accessed   electronically from a computer system. &
      A system for storing and taking care of data &
      3.67 \\ \hline
    23 &
      An organized collection of data, generally stored and accessed   electronically from a computer system. &
      A set of rules and methods that describe the functionality, organization,   and implementation of computer systems. &
      1.60 \\ \hline
    24 &
      A system for storing and taking care of data &
      A large collection of data organized especially for rapid   search and retrieval &
      2.87 \\ \hline
    25 &
      A set of rules and methods that describe the functionality, organization,   and implementation of computer systems. &
      The conceptual design and fundamental operational structure of   a computer system. &
      3.27 \\ \hline
    26 &
      A large collection of data organized especially for rapid   search and retrieval &
      The manner in which the components of a computer or computer system are   organized and integrated &
      1.13 \\ \hline
    27 &
      The conceptual design and fundamental operational structure of   a computer system. &
      The manner in which the components of a computer or computer system are   organized and integrated &
      3.93 \\ \hline
    28 &
      An informal high-level description of the operating principle   of a computer program or other algorithm. &
      A description of the source code of a computer program or an   algorithm in a language easily understood by humans. &
      3.93 \\ \hline
    29 &
      An informal high-level description of the operating principle   of a computer program or other algorithm. &
      a formal language comprising a set of   instructions that produce various kinds of output. &
      2.07 \\ \hline
    30 &
      A description of the source code of a computer program or an   algorithm in a language easily understood by humans. &
      A program code unrelated to the hardware of a particular computer and   requiring conversion to the code used by the computer before the program can   be used. &
      2.80 \\ \hline
    31 &
      a formal language comprising a set of   instructions that produce various kinds of output. &
      a type of written language that tells computers what   to do in order to work. &
      3.60 \\ \hline
    32 &
      A program code unrelated to the hardware of a particular computer and   requiring conversion to the code used by the computer before the program can   be used. &
      various high-level languages used for computer programs &
      2.33 \\ \hline
    33 &
      a type of written language that tells computers what   to do in order to work. &
      various high-level languages used for computer programs &
      3.80 \\ \hline
    34 &
      The discovery, interpretation, and communication of meaningful patterns   in data. &
      The process of discovering, interpreting, and communicating significant   patterns in data to get meaningful information. &
      4.27 \\ \hline
    35 &
      The discovery, interpretation, and communication of meaningful patterns   in data. &
      A branch of information technology which is intended to   protect computers. &
      0.60 \\ \hline
    36 &
      The protection of computer systems and networks from   the theft of or damage to their hardware, software,   or electronic data. &
      A branch of information technology which is intended to   protect computers. &
      3.73 \\ \hline
    37 &
      The protection of computer systems and networks from   the theft of or damage to their hardware, software,   or electronic data. &
      The process of discovering, interpreting, and communicating significant   patterns in data to get meaningful information. &
      0.33 \\ \hline
    38 &
      A type of computer program that, when executed, replicates   itself by modifying other computer programs and inserting its own code. &
      A program that is able to copy itself when it is run and can also  execute instructions that cause harm. &
      4.27 \\ \hline
    39 &
      The practice of storing regularly used computer data on multiple servers   that can be accessed through the Internet &
      A computer program that is usually disguised as an innocuous program or   file, that often produces copies of itself and inserts them into other   programs performing a malicious action &
      0.13 \\ \hline
    40 &
      The on-demand availability of computer system resources,   especially data storage and computing power, without direct active   management by the user. &
      Computing services provided by a company or   place outside of where they are being used. &
      2.73 \\ \hline
    41 &
      The on-demand availability of computer system resources,   especially data storage (cloud storage) and computing power, without   direct active management by the user. &
      The practice of storing regularly used computer data on multiple servers   that can be accessed through the Internet &
      2.60 \\ \hline
    42 &
      Computing services provided by a company or   place outside of where they are being used. &
      A type of computer program that, when executed, replicates   itself by modifying other computer programs and inserting its own code. &
      0.47 \\ \hline
    43 &
      A program that is able to copy itself when it is run and can also  execute instructions that cause harm. &
      A computer program that is usually disguised as an innocuous program or   file, that often produces copies of itself and inserts them into other   programs performing a malicious action &
      4.33 \\ \hline
    44 &
      A piece of computer hardware or software that provides functionality for   other programs or devices, called "clients". &
      A computer that serves many kinds of information to   a user or client machine. &
      3.60 \\ \hline
    45 &
      A piece of computer hardware or software that provides functionality for   other programs or devices, called "clients". &
      A computer in a network that is used to provide services to   other computers in the network &
      2.87 \\ \hline
    46 &
      A network security system that monitors and controls   incoming and outgoing network traffic based on predetermined   security rules. &
      A piece of software that monitors the network traffic between the   inside and outside. &
      4.20 \\ \hline
    47 &
      A network security system that monitors and controls   incoming and outgoing network traffic based on predetermined   security rules. &
      A computer hardware or software that prevents unauthorized access to   private data by outside computer users &
      3.40 \\ \hline
    48 &
      A computer that serves many kinds of information to   a user or client machine. &
      A piece of software that monitors the network traffic between the   inside and outside. &
      1.20 \\ \hline
    49 &
      A computer in a network that is used to provide services to   other computers in the network &
      A computer hardware or software that prevents unauthorized access to   private data by outside computer users &
      0.87 \\ \hline
    50 &
      A data point that differs significantly from other observations. &
      A statistical observation that is markedly different in value from the   others of the sample &
      4.13 \label{tab:datasettable}\\ \hline
    \caption{Complex Sentence Dataset } 
    \end{longtable}
\end{footnotesize}
\clearpage
\section{Survey environments} \label{appendix: B}
\begin{figure}[h]
  \centering
  \begin{minipage}[b]{0.45\textwidth}
    \includegraphics[width=\textwidth]{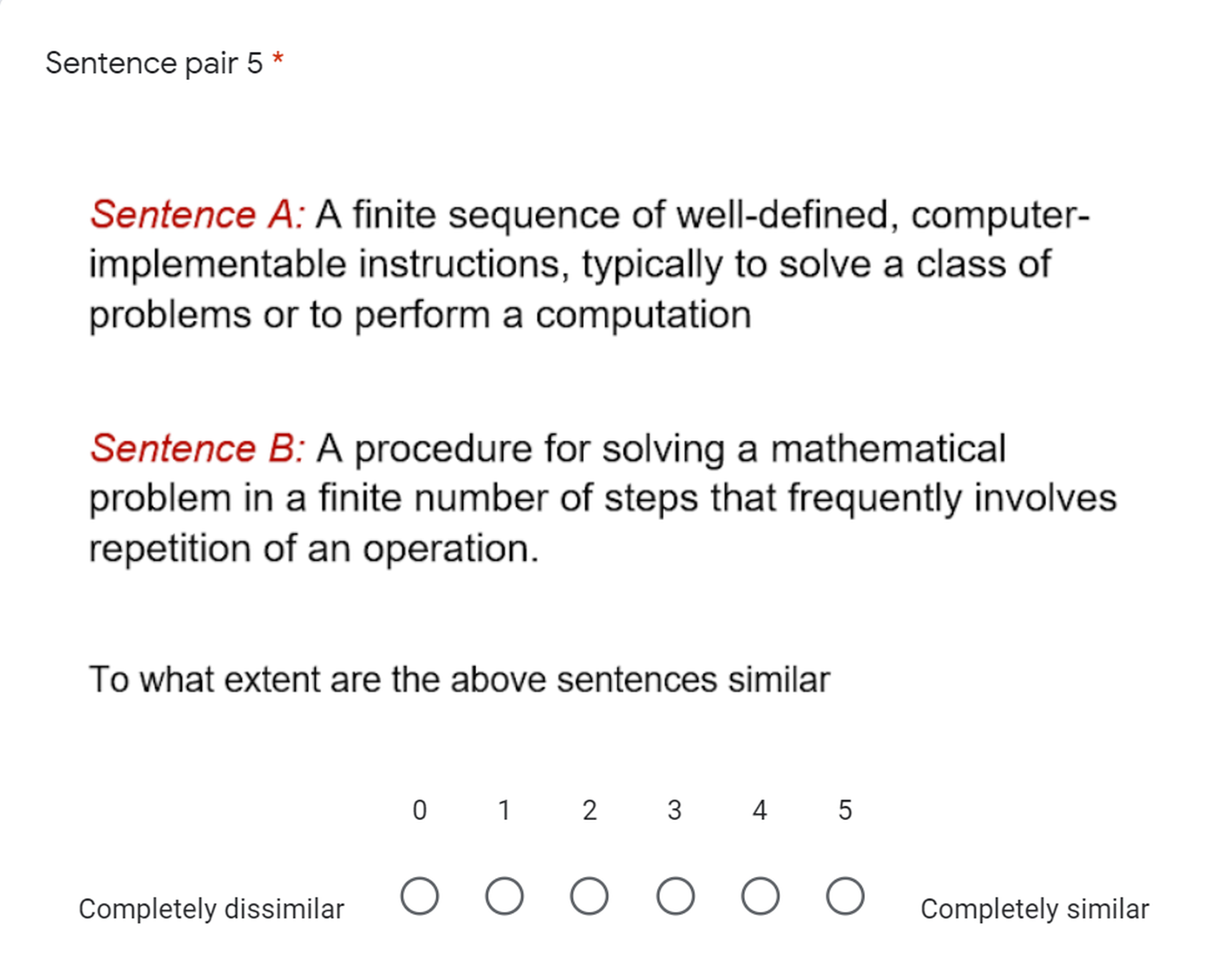}
    \caption{Survey using Google forms}
    \label{FIG: F5}
  \end{minipage}
  \hfill
  \begin{minipage}[b]{0.45\textwidth}
    \includegraphics[width=\textwidth]{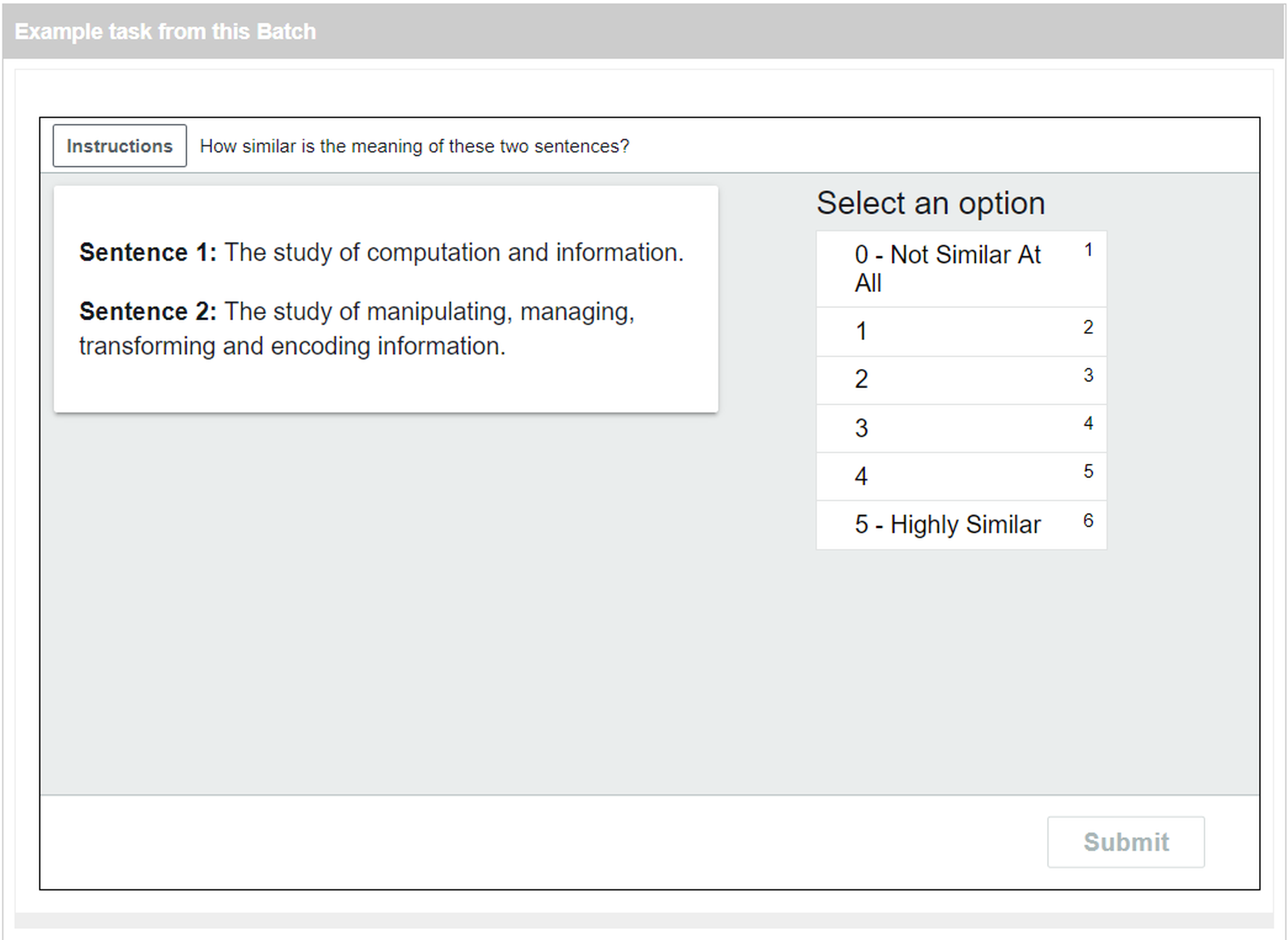}
    \caption{Survey using Amazon Mechanical Turk}
    \label{FIG: F6}
  \end{minipage}
\end{figure}

\section{Semantic similarity Datasets}\label{appendix: C}
\begin{table}[h]
  \caption{Popular Benchmark datasets for Semantic similarity\cite{chandrasekaran2021evolution}}
    \label{tab:Dataset list}
\begin{tabularx}{\textwidth}{p{4cm}|p{4cm}|p{3cm}|p{ 1cm}|p{1.5cm}}
  \toprule
    \textbf{Dataset Name} & \textbf{Word/Sentence pairs} & \textbf{Similarity score range} & \textbf{Year} & \textbf{Reference} \\ \midrule
    R\&G       & 65    & 0-4  & 1965 & \cite{rubenstein1965contextual} \\ 
    M\&C       & 30    & 0-4  & 1991 & \cite{miller1991contextual} \\ 
    WS353      & 353   & 0-10 & 2002 & \cite{finkelstein2001placing}\\ 
    LiSent     & 65    & 0-4  & 2007 &  \cite{li2006sentence}\\ 
    SRS        & 30    & 0-4  & 2007 &  \cite{pedersen2007measures}\\ 
    WS353-Sim  & 203   & 0-10 & 2009 & \cite{agirre2009study} \\
    STS2012 & 5250   & 0-5  & 2012 &  \cite{agirre2012semeval} \\
    STS2013 & 2250   & 0-5  & 2013 &  \cite{agirre2013sem} \\ 
    WP300      & 300   & 0-1  & 2013 &  \cite{li2013computing} \\ 
    STS2014 & 3750   & 0-5  & 2014 &  \cite{agirre2014semeval} \\ 
    SL7576     & 7576  & 1-5  & 2014 &  \cite{silberer2014learning}\\ 
    SimLex-999 & 999   & 0-10 & 2014 &  \cite{hill2015simlex}\\ 
    SICK       & 10000 & 1-5  & 2014 &  \cite{marelli2014sick}\\ 
    STS2015 & 3000   & 0-5  & 2015 &  \cite{agirre2015semeval} \\ 
    SimVerb    & 3500  & 0-10 & 2016 &  \cite{gerz2016simverb} \\ 
    STS2016 & 1186   & 0-5  & 2016 &  \cite{agirre2016semeval} \\ 
    WiC        & 5428  & NA   & 2019 &  \cite{pilehvar2019wic}\\ \bottomrule
   
    \end{tabularx}
\end{table}
\EOD
\end{document}